\documentclass[final]{cvpr}

\usepackage{times}
\usepackage{epsfig}
\usepackage{graphicx}
\usepackage{amsmath}
\usepackage{amssymb}

\usepackage{algorithmic}
\usepackage{algorithm}
\usepackage{booktabs}
\newcommand{\bhline}[1]{\noalign{\hrule height #1}}
\newcommand{\bm}[1]{{\mbox{\boldmath $#1$}}} 
\usepackage[subrefformat=parens]{subcaption}
\usepackage[pagebackref=true,breaklinks=true,colorlinks,bookmarks=false]{hyperref}

\def\cvprPaperID{23} 
\def\confYear{CVPR 2021}
\makeatletter
\def\thanks#1{\protected@xdef\@thanks{\@thanks
        \protect\footnotetext{#1}}}
\makeatother

\begin{document}

\title{Multi-task Learning with Attention for End-to-end Autonomous Driving}


\author{
    Keishi Ishihara$^{1, 2}$, Anssi Kanervisto$^2$, Jun Miura$^1$, Ville Hautam\"aki$^2$\\
    $^1$Toyohashi University of Technology\\
    $^2$University of Eastern Finland\\
    {\tt\small ishihara@aisl.cs.tut.ac.jp, jun.miura@tut.jp, \{anssk, villeh\}@cs.uef.fi}
    
    \thanks{
    \textcopyright 2021 IEEE.  Personal use of this material is permitted.  Permission from IEEE must be obtained for all other uses, in any current or future media, including reprinting/republishing this material for advertising or promotional purposes, creating new collective works, for resale or redistribution to servers or lists, or reuse of any copyrighted component of this work in other works.
    }
}

\maketitle
\thispagestyle{empty}

\begin{abstract}
 Autonomous driving systems need to handle complex scenarios
 such as lane following, avoiding collisions, taking turns,
 and responding to traffic signals. In recent years, approaches
 based on end-to-end behavioral cloning have demonstrated
 remarkable performance in point-to-point navigational scenarios,
 using a realistic simulator and standard benchmarks. Offline
 imitation learning is readily available, as it does not require
 expensive hand annotation or interaction with the target
 environment, but it is difficult to obtain a reliable system.
 In addition, existing methods have not specifically addressed
 the learning of reaction for traffic lights, which are a rare
 occurrence in the training datasets. Inspired by the previous work
 on multi-task learning and attention modeling, we propose
 a novel multi-task attention-aware network in the conditional
 imitation learning (CIL) framework. This does not only improve
 the success rate of standard benchmarks, but also the ability
 to react to traffic lights, which we show with standard benchmarks.
\end{abstract}

\section{\label{sec:introduction} Introduction}


\begin{figure}[!h]
\begin{center}
\includegraphics[width=1.\linewidth]{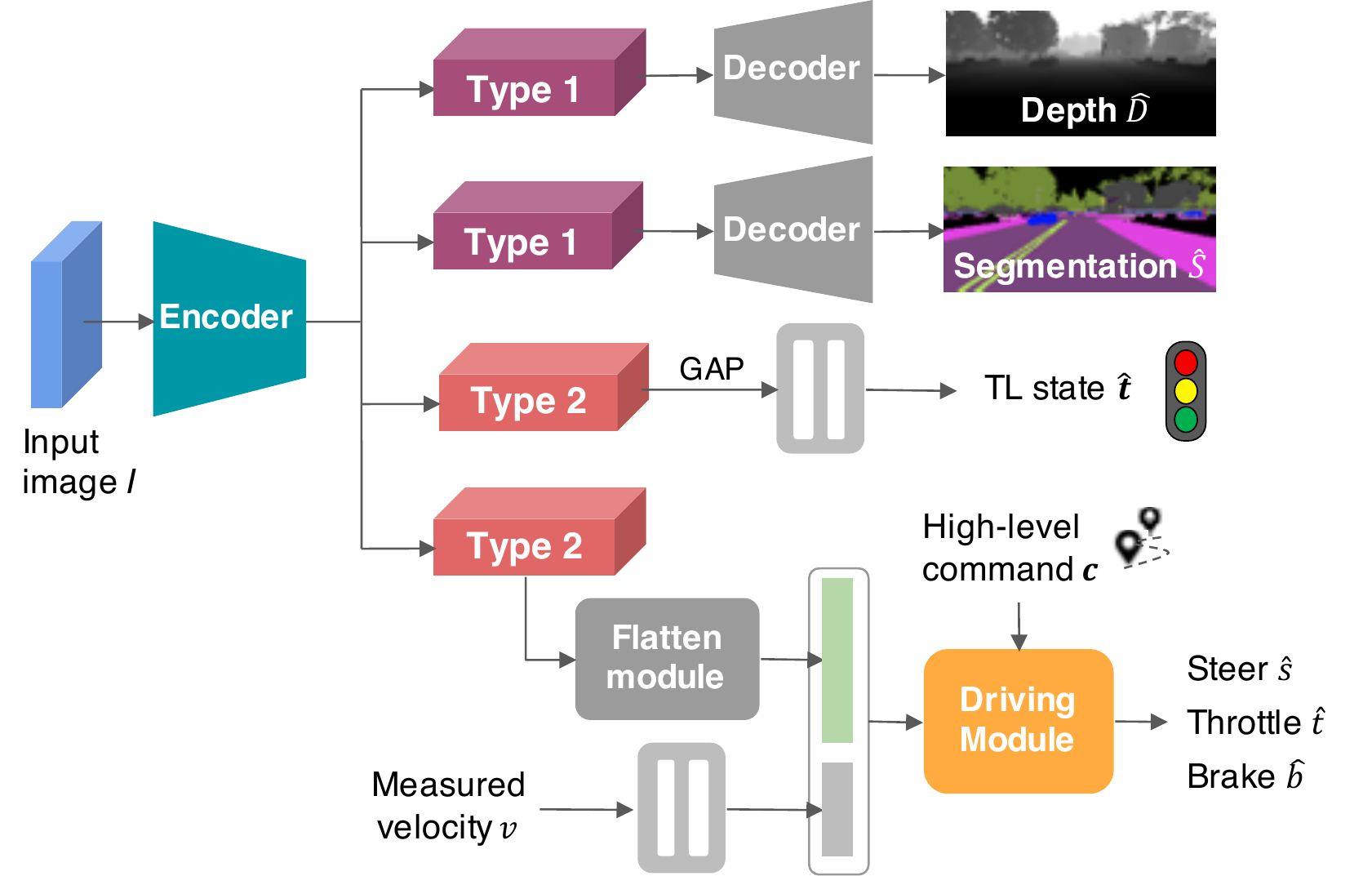}
\end{center}
  \caption{
  An overview of our proposed multi-task attention-aware network architecture which is composed of seven components: encoder, two decoders, traffic light (TL) state classifier, flatten module, velocity encoder, and driving module. The main target task of this network is to predict control signals from a monocular RGB camera end-to-end. The encoder implements an attention mechanism and generates two types of attention-weighted latent feature maps (see Figure \ref{fig:encoder}).}
\label{fig:overview}
\end{figure}

In the field of autonomous driving, end-to-end behavioral
cloning has emerged recently, and many deep networks trained
to mimic expert demonstrations have shown reasonable
performances from lane-keeping \cite{bib:bojarski2016,
bib:xu2017end} to point-to-point navigation
\cite{bib:cil2018, bib:codevilla2019}.
Typically, those deep networks are trained by using an offline
dataset generated by an expert-driver (off-policy), or they
lean a control policy by rolling out the environment itself
to find good rewards using reinforcement learning (on-policy).

However, previous studies have revealed several critical limitations.
Firstly, an agent that learned a control policy from
an offline dataset may not be able to make accurate decisions
in previously unobserved environments.
In a study \cite{bib:codevilla2019} conducted by using an
open-source driving simulator CARLA \cite{bib:carlasim2017},
it was reported that the driving performance of the imitation
learning agent considerably drops under those
conditions such as untrained urban area, weather conditions,
and traffic congestion. Secondly, it is important to know
how well a network perceives visual inputs for such a
safety-critical application of autonomous driving,
but only a few studies
addressed this issue \cite{bib:VisualAtt2020, bib:rethinking2018,
bib:ABN_for_IL_2019}.

Therefore, in this work, we tackle the aforementioned problems
based on the following ideas:
(1) Learning visual recognition sub-tasks (e.g., semantic segmentation)
alongside can encourage the network to learn generalizable
scene representations that are effective for decision-making.
Similarly to \cite{bib:rethinking2018}, we also introduce
multi-tasking to decode scene representations as sub-tasks.
Although, in the previous study, the learning process was divided
into perception and control, we train all tasks at the same time.
Additionally, we adopt traffic light classification as one of the sub-tasks.
In this way, an operator can easily debug the model performance
by looking at its predictions. (2) Also, an attention mechanism
is introduced to further improve control performance by focusing
on salient regions among the extracted features like humans do when driving.
Our main contributions can be summarized as follows:
\begin{itemize}
  \item We propose a novel multi-task attention-aware network
  for vision-based end-to-end autonomous driving.
  \item We show that our approach achieves better or comparable
  results with the current state-of-the-art models on the CARLA
  benchmarks \cite{bib:codevilla2019, bib:carlasim2017}.
  \item We further perform the traffic light infraction
  analysis to quantitatively show the effectiveness of
  our approach in handling traffic lights.
  \item We study how attention layers change what network
  focuses on by visualizing saliency maps.
\end{itemize}

\section{\label{sec:related_work} Related work}

\paragraph{Behavioral cloning for autonomous driving}

Bojarski \etal \cite{bib:bojarski2016} were the first to 
successfully demonstrate lane following task in an end-to-end
(image-to-steering) manner using a simple CNN. 
Later, Xu \etal \cite{bib:xu2017end} used a large-scale video dataset
to predict vehicle egomotion.
To extend the learned control policy to goal-directed 
navigation to solve the ambiguous action problem (e.g., intersections),
Codevilla \etal \cite{bib:cil2018} proposed Conditional Imitation
Learning (CIL) framework. The goal of CIL is to
navigate in an urban environment where the autonomous vehicle
must take turns at intersections based on high-level command
input such as ``turn left'', ``turn right'', and ``go straight''.
Because of CIL and open-source access to realistic driving simulator
CARLA \cite{bib:carlasim2017}, many follow-up works have been conducted
in CIL framework \cite{bib:lbc2019, bib:codevilla2019, bib:VisualAtt2020,
bib:rethinking2018, bib:CIRL2018,  bib:CAL2018} which called
a vision-based driving system where perception relies on visual input
from cameras only to keep the entire system simple.

\paragraph{Attention in vision models}

In the field of computer vision, attention has been a key idea to
improve performance of CNNs in various tasks such as classification 
\cite{bib:SENet2018, bib:CBAM2018}, object detection \cite{bib:fan2020few},
image tracking \cite{bib:imagetracking2012},
and captioning~\cite{bib:showtell2015}.
Katharopoulos and Fleuret \cite{bib:megapixel_attention2019} proposed 
an attention-sampling approach to process megapixel images .
Liu \etal.~\cite{bib:MTAN2019} presented the Multi-Task Attention Network
(MTAN) that has a feature-level attention mechanism to select task-specific
features for multi-task learning.
Usually, incorporating an attention mechanism requires a unique architectural
design like the abovementioned ones. In contrast, module-based attention
approaches \cite{bib:SENet2018, bib:RAN2017, bib:CBAM2018} are another
trend that can simply be attached to existing CNNs
\cite{bib:Xception_CVPR2017, bib:ResNet_CVPR2016, bib:MobileNet_2017}
and improves performances.
Convolutional Block Attention Module (CBAM) \cite{bib:CBAM2018} is
a representative of them, where attention is learnable end-to-end
without requiring additional labels.
A few approaches have also introduced attention in the autonomous driving context.
Mori \etal \cite{bib:ABN_for_IL_2019} used Attention Branched Network 
\cite{bib:ABNCVPR2019} to develop a visually interpretable model.
Similarly, Kim and Canny \cite{bib:kim2017interpretable} visualized 
saliency part to predict steering by decoding attention heat maps.

An approach similar to ours by Cultrera \etal \cite{bib:VisualAtt2020}
splits features into multi-scale grids and weighs with softmax scores to
drop out irrelevant regions.
In comparison, our approach uses CBAM specialized for one particular task
in multi-task architecture to emphasize features over the channel and
spatial dimensions separately and refines specifically to the task.
Also, our attention mechanism is based on MTAN \cite{bib:MTAN2019},
which originally handles dense estimation tasks only such as
semantic segmentation and depth estimation, while we design it to fit into
the CIL framework \cite{bib:cil2018} and our model performs control prediction
in an end-to-end manner as well.

\section{\label{sec:method} Methodology}

\begin{figure*}[h!]
\begin{center}
\includegraphics[width=1.\linewidth]{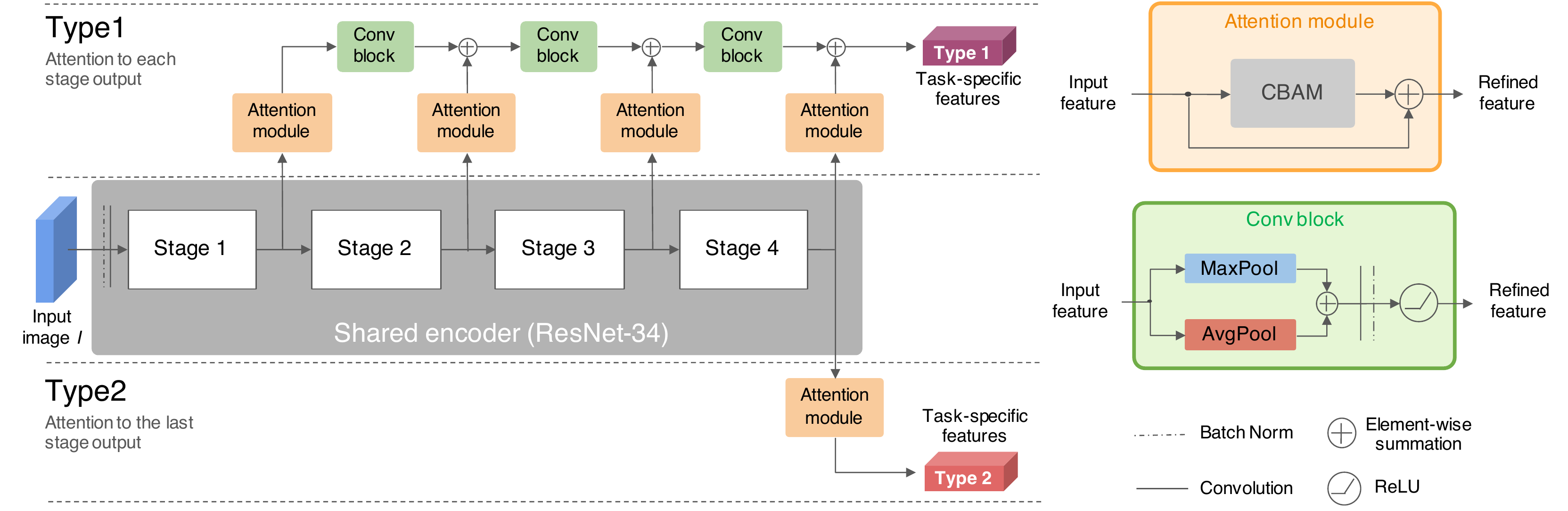}
\end{center}
  \caption{
  Visualization of the ResNet34-based encoder network with two attention mechanisms. This encoder outputs two types of task-specific attention-weighted feature maps: $Type1$ and $Type2$, generated by its attention paths formed by attention modules and convolution blocks, similar to MTAN \cite{bib:MTAN2019}. 
  }
\label{fig:encoder}
\end{figure*}

Inspired by the previous works of end-to-end driving 
\cite{bib:cil2018, bib:rethinking2018} and visual
attention methods \cite{bib:MTAN2019, bib:CBAM2018},
we introduce a multi-task attention-aware neural
network that learns compact road scene representation to
jointly predict scene representations (semantic
segmentation and depth estimation), color state of
traffic light, and driving control.

First of all, we consider the autonomous driving problem in
an urban environment as a goal-directed motion planning task.
To this end, we adopt the Conditional Imitation Learning (CIL)
framework \cite{bib:cil2018} to train our model in an end-to-end manner.

Secondly, our approach learns semantic segmentation and depth
estimation to learn to encode the image into a meaningful
feature vector as in \cite{bib:rethinking2018}. Although the
authors of \cite{bib:rethinking2018} were the first to introduce
those scene representation learning in the CIL framework,
the learning process was divided into perception and driving.
In our approach, we train segmentation and depth estimation
together with control.
Moreover, we also add traffic light classification task as
the fourth task that our network performs at the same time
as the others.

Thirdly, our approach incorporates an attention mechanism
inspired by the MTAN proposed by Liu \etal in~\cite{bib:MTAN2019}
to successfully learn task-shared and task-specific features separately.
Differently from the original MTAN which drops irrelevant features
from global feature maps, we use a CBAM-based attention module to
emphasize the global feature maps by additive operation.
This contributes to the learning of much more complex behavior
of reacting to traffic lights than the ability to keep
lane and following vehicles.


\subsection{\label{sec:network} Network architecture}

Figure \ref{fig:overview} gives an overview of the architectural
design of our proposed approach which is composed of seven
components: encoder, two decoders for scene representations,
traffic light state classifier, flatten module, velocity encoder,
and driving module.
Our network model is built on the two state-of-the-art
approaches in the CARLA benchmark;
the Conditional Imitation Learning approach with ResNet-34
backbone and Speed prediction head (CILRS) proposed by Codevilla
\etal in \cite{bib:codevilla2019}, and the Multi-Task learning
approach (MT) proposed by Li \etal in \cite{bib:rethinking2018}.

In the CIL framework, the agent is conditioned by the
external navigational command (e.g., turn left) and
asked for the control command corresponding to it 
so that the learned agent can be controlled by that
command at test time. This eliminates the ambiguous
problem at intersections where multiple actions can be
taken \cite{bib:cil2018}.

Our model's inputs are a monocular RGB camera image $I$ from a front-facing camera,
a measured velocity $v$ of the ego-vehicle, and a high-level command ${\bm c}$ which is
a one-hot encoded vector of (``follow lane'', ``turn left'', ``turn right'',
``go straight''). Our model performs predictions of control signals (steering
$\hat{s}$, throttle $\hat{t}$, and brake $\hat{b}$) as main task along with three
different complementary sub-tasks: semantic segmentation $\hat{S}$, depth
estimation $\hat{D}$, and classification of the traffic light state
$\hat{\bm t}$ which is one-hot vector of (``red'',
``yellow``,``green'', ``none'') at every frame.
By doing such sub-tasks, we suppose that the model can capture 
road scenes without missing necessary information for
generating control signals and encode those into its latent feature maps. 
Additionally, we employ a simple channel-spatial attention mechanism 
described in the next paragraph inspired by the study \cite{bib:MTAN2019}
where task-specific attention modules are used to
select and emphasize features from one shared encoder for different image
recognition tasks.

\paragraph{Residual network with attention}
\label{par:encoder}

Figure \ref{fig:encoder} shows the encoder module
that consists of ResNet-34 \cite{bib:ResNet_CVPR2016} for 
the backbone network,
attention modules, and convolutional blocks.
Using these attention modules and convolution blocks,
we build two types
of task-specific attention paths along with the
shared backbone encoder: $Type1$ and $Type2$.
$Type 1$ is formed by a set of attention 
modules and convolution blocks, which then takes
convolution feature maps from every end of the 
ResNet stages.
The attention modules play as feature refiners that 
emphasize the global feature maps
specifically to a certain task.
In comparison, MTAN \cite{bib:MTAN2019} used attention
to select which features to use, while we emphasize them via
an additive operation.


The $Type 2$ is made of an attention module that
refines the feature
maps obtained from the last layer of the ResNet.
In total, we prepare two $Type 1$ attention paths
for segmentation and depth estimation, and two
$Type 2$ paths for traffic light classification
and control command prediction.

We have empirically found that combining the two types
of attention mechanisms results in better performance
than using only $Type 1$ or $Type 2$ individually for all tasks.
Intuitively this is reasonable because tasks that require
dense estimation, such as segmentation and depth
estimation, may often be better to have access to 
higher resolution feature maps to avoid missing details,
as is a common technique in architectures like U-Net
\cite{bib:UNET2015}. Conversely, tasks such as control 
prediction and traffic light classification may require
more abstract features in the hidden 
representation.
For the attention module, we adopt CBAM \cite{bib:CBAM2018}.

\paragraph{Decoder networks}
As shown in Figure \ref{fig:overview}, we build two decoder
networks for the image decoding sub-tasks: semantic segmentation
and depth estimation, each receiving
$Type 1$ task-specific features from the encoder.
The goal of those sub-tasks is to learn compact and generalizable
latent representation for control prediction as done in
\cite{bib:rethinking2018, bib:xu2017end}.
This multi-tasking approach enforces the encoder to generate
representations such that they encode ``what'', ``where'',
and ``how far'' information in the feature maps.
The resolution of both decoded images is the same as the input
image resolution of $384 \times 160$.
The decoder architectures are based on \cite{bib:rethinking2018},
but empirically we have found that performance is improved
when the segmentation decoder larger than the depth one.

\paragraph{Traffic light classifier}
Reacting to the traffic lights has not yet been explicitly
modeled in vision-based end-to-end driving approaches
\cite{bib:cil2018, bib:VisualAtt2020,  bib:rethinking2018}.
In some works \cite{bib:lbc2019, bib:codevilla2019}, 
the authors addressed the learning of the behavior
of reacting to the traffic lights from only the demonstration
of stopping for a red light and starting going forward when
it turns green. However, in situations where an agent is 
facing a traffic light, it is important for humans
to know which state of the traffic light the agent
is aware of. Also, the ability to classify the light state
can help the network learn the behavior of reacting
to it according to the color. Therefore, we introduce
traffic light state classification as one of the sub-tasks.
Our approach classifies a traffic light state at every frame into
one of four classes including (``red'', ``yellow'', ``green'', ``none'').

\paragraph{Driving module}
The driving module takes a vector feature map denoted
as $j$ produced by concatenating flattened feature map, encoded velocity input $v$,
high-level command $\bm{c}$ and outputs control
commands (steering $\hat{s}$, throttle $\hat{t}$ and
braking signal $\hat{b}$), which are referred to as
regression tasks.
Following prior works in the CIL framework
\cite{bib:lbc2019, bib:cil2018, bib:codevilla2019,
bib:VisualAtt2020, bib:rethinking2018,bib:CAL2018},
our driving module also implements branched prediction head:
four branches corresponding to each navigational command.
Command input selects which of these branches is used
to predict control commands.
The agent that has the command-dedicated prediction
head performs better than that with one prediction
head, which is reported by Codevilla \etal in \cite{bib:cil2018}.

\subsection{The model objective}
\label{sec:losses}

To the network, we define the objective function $\mathcal{L}_{total}$ as the weighted sum of control regression loss $\mathcal{L}_{control}$, traffic light state classification loss $\mathcal{L}_{tl}$, semantic segmentation loss $\mathcal{L}_{seg}$, and depth estimation loss $\mathcal{L}_{dep}$ with the task weightings denoted by $\lambda$ with each task subscript:
\begin{equation}
    \mathcal{L}_{total}=\lambda_{control}\mathcal{L}_{control}+\lambda_{tl}\mathcal{L}_{tl}+\lambda_{seg}\mathcal{L}_{seg}+\lambda_{dep}\mathcal{L}_{dep}
    \label{eq:loss_total}
\end{equation}
Control loss $\mathcal{L}_{control}$ is the weighted linear combination
of steering, throttle, and brake losses $\mathcal{L}_{c}$ with weighting
terms $\gamma_{c}$, where $c=1$ is for steering, $c=2$ for throttle,
and $c=3$ for brake:
\begin{equation}
    \mathcal{L}_{control}=\sum_{c=1}^{3} \gamma_{c}\mathcal{L}_{c}
    \label{eq:loss_control}
\end{equation}
We use a mean squared error for each control regression loss $\mathcal{L}_{c}$. 
For traffic light state classification and semantic segmentation,
we use class-weighted categorical cross-entropy losses: $\mathcal{L}_{seg}$ calculated between the network output $\hat{S}$ and ground-truth label $S$, and $\mathcal{L}_{tl}$ done between $\hat{\bm{t}}$ and $\bm{t}$.
Lastly, for depth estimation, we use a pixel-wise mean 
squared error between the network output
$\hat{D}$ and the ground-truth label $D$.
All coefficients $\lambda$ and $\gamma$ are obtained 
empirically by examining validation errors to determine
when they were valid.

\begin{figure}[!t]
    \begin{center}
        \includegraphics[width=1.0\linewidth]{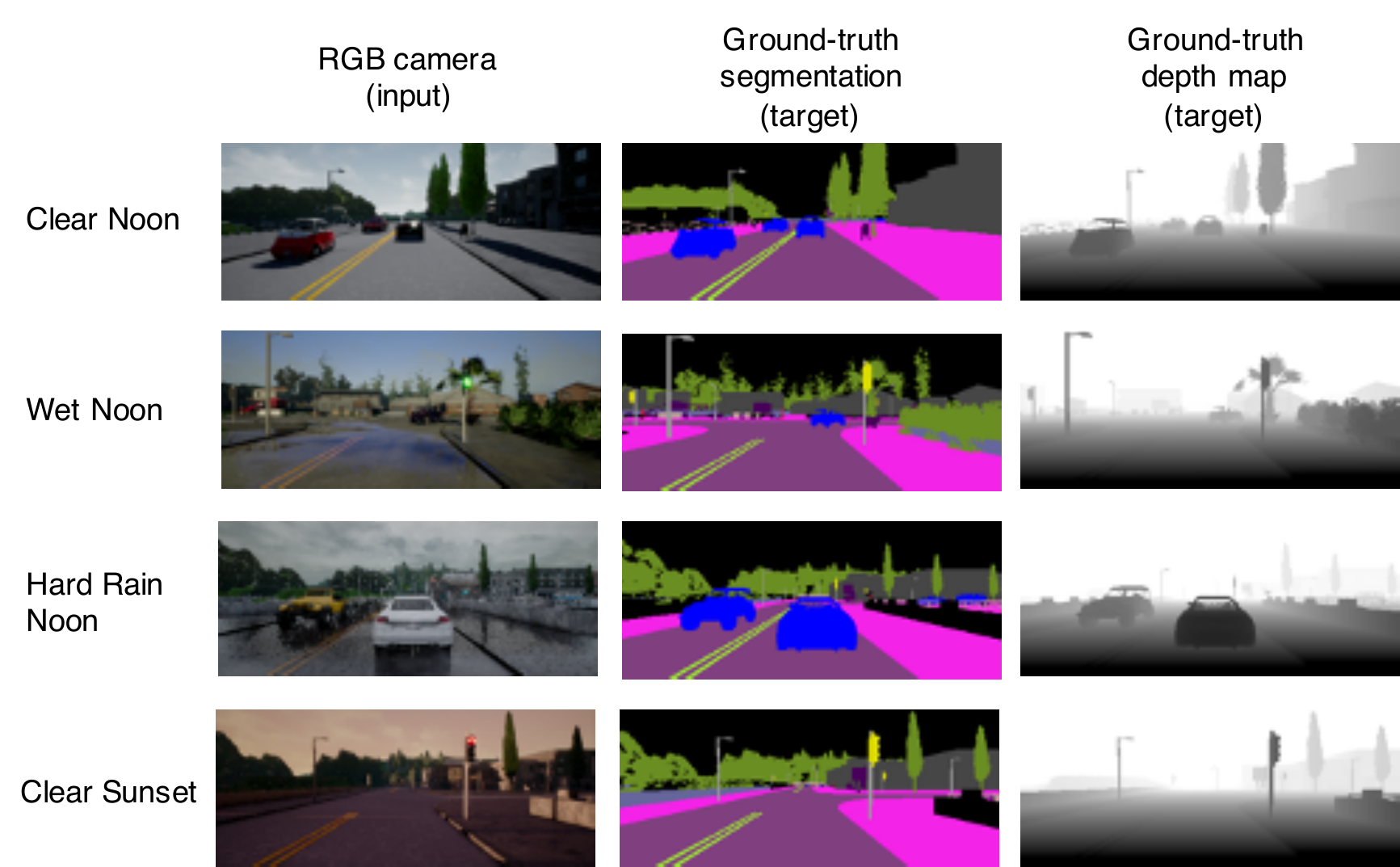}
        \subcaption{Training weather conditions in Town01}
        \includegraphics[width=1.0\linewidth]{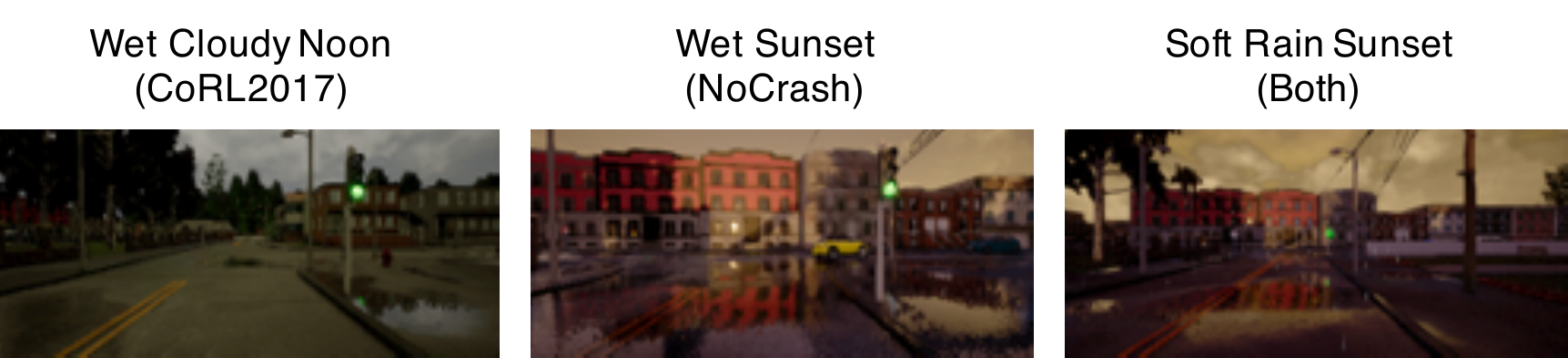}
        \subcaption{New weather conditions in Town02}
    \end{center}
    \caption{\label{fig:data_examples}
    Example frames of the collected dataset (training conditions in Town01) and new weather conditions in Town02.}
\end{figure}

\section{Environment and dataset}
\label{sec:dataset}

We perform all our experiments in CARLA simulator~\cite{bib:carlasim2017} version 0.8.4.
This version of the simulator provides two different
suburban towns: Town01 for training, and Town02 for testing.
It has a diverse, non-deterministic environment
where dynamic obstacles may appear (e.g auto-cruising
vehicles and pedestrians) and weather conditions can 
also change by user preferences.
In particular, pedestrians have random behaviors;
crossing the street without any previous notice, 
which is unpredictable and thus an agent
needs to react quickly as described in 
\cite{bib:codevilla2019}. 
Moreover, they may crash themselves into a car 
stopping at a red light or an obstacle, resulting in 
an incomplete evaluation.

\subsection{Dataset collection}

We collected our dataset simply because provided one by 
Codevilla \etal in \cite{bib:codevilla2019} do not
contain traffic light information.
To do so, we adapt a hand-coded expert autopilot
provided by Codevilla \etal in \cite{bib:cil2018}, 
which leverages simulated information to follow
the waypoints nearly perfectly and stop for
forward-driving vehicles and pedestrians to
avoid collisions.
It also reacts to the traffic lights regarding
their color state. 
To obtain the ground-truth label of the light state,
the original code provided was modified to save it 
additionally.
When multiple lights are visible, the state of the
light facing the ego-vehicle is saved as the 
ground-truth label.
We collected $466,000$ frames from Town01 and
randomly split into $372,000$ of the training
set and $94,000$ of the validation set.
Each frame consists of front-faced monocular RGB 
camera image $I$,
auto-generated semantic segmentation $S$,
depth map $D$, and measurements including
steering $s$, throttle $t$, brake $b$, speed $v$, 
high level command $\bm{c}$, and ground-truth
traffic light state $\bm{t}$.
The dataset is collected under four different
weather conditions (``Clear Noon'', ``Wet Noon'',
``Hard Rain Noon'', ``Clear Sunset'') as it is
specified condition by the CARLA benchmarks,
described more in Section \ref{sec:experimental_setup}.

\subsection{Data augmentation and balancing}

We found that data augmentation and balancing are
crucial for better generalization performance as
reported in previous works \cite{bib:cil2018,
bib:rethinking2018}.
In order to enlarge training dataset distribution,
we use the same set of common image augmentations
as done in \cite{bib:cil2018}, including gaussian noise,
blurring, pixel dropout, and contrast
normalization during training.
Furthermore, those image transformations are followed by
PCA color augmentation which is proposed in
\cite{bib:AlexNet2012} across every mini-batch.

Additionally, it is well known that driving datasets
have inherently unbalanced distribution in general,
especially in control \cite{bib:amini2018variational}.
Our dataset is no exception and thus contains mostly 
driving forwards, so without balancing agent only drives
forward. To ease the bias of the skewed distribution of our dataset,
we performed undersampling based on steering values during
training. 
Specifically, a sample has a smaller chance of being selected
if it is overrepresented in the dataset. 
Consequently, straight driving scenes which are the dominant
mode in the dataset are often skipped when training.
Without this balancing, our agents rarely completed full
episodes in the benchmarks.

\begin{table}
\begin{center}
\setlength\tabcolsep{4pt}
\begin{tabular}{lcc|cc}
\hline
             & \multicolumn{4}{c}{Training conditions (4 weathers, Town01)}                                    \\
Task         & MT                     & CILRS               & Ours w/o Att.         & Ours                     \\ \hline
Straight     &  \textbf{100} $\pm$ 0  & 98 $\pm$ 3          & 99 $\pm$ 1            &  \textbf{100} $\pm$ 0    \\
One Turn     &  \textbf{100} $\pm$ 1  & 92 $\pm$ 7          & 97 $\pm$ 4            &  \textbf{100} $\pm$ 1    \\
Navigation   &  \textbf{100} $\pm$ 0  & 97 $\pm$ 4          & 98 $\pm$ 3            &  \textbf{100} $\pm$ 0    \\
Nav. Dyn.    &  \textbf{99} $\pm$ 3   & 95 $\pm$ 4          & 96 $\pm$ 4            &  \textbf{99} $\pm$ 2     \\ \hline\hline
             & \multicolumn{4}{c}{New weather (2 weathers, Town01)}                                            \\
Task         & MT                     & CILRS               & Ours w/o Att.         & Ours                     \\ \hline
Straight     & \textbf{100} $\pm$ 0   & 95 $\pm$ 5          & 98 $\pm$ 4            & \textbf{100} $\pm$ 0     \\
One Turn     & \textbf{100} $\pm$ 1   & 85 $\pm$ 8          & 94 $\pm$ 7            & 99 $\pm$ 3               \\
Navigation   & \textbf{100} $\pm$ 0   & 71 $\pm$ 6          & 94 $\pm$ 9            & 97 $\pm$ 6               \\
Nav. Dyn.    & \textbf{99} $\pm$ 3    & 70 $\pm$ 4          & 96 $\pm$ 7            & 97 $\pm$ 4               \\ \hline\hline
             & \multicolumn{4}{c}{New town (4 weathers, Town02)}                                               \\
Task         & MT                     & CILRS               & Ours w/o Att.         & Ours                     \\ \hline
Straight     & 98 $\pm$ 4             & 95 $\pm$ 4          & 98 $\pm$ 4            & \textbf{99 $\pm$ 3}      \\
One Turn     & 93 $\pm$ 6             & 85 $\pm$ 7          & 92 $\pm$ 8            & \textbf{98 $\pm$ 2}      \\
Navigation   & 81 $\pm$ 12            & 71 $\pm$ 9          & 84 $\pm$ 14           & \textbf{93 $\pm$ 10}     \\
Nav. Dyn.    & 78 $\pm$ 16            & 70 $\pm$ 14         & 82 $\pm$ 13           & \textbf{91 $\pm$ 11}     \\ \hline\hline
             & \multicolumn{4}{c}{New town \& weather (2 weathers, Town02)}                                    \\
Task         & MT                     & CILRS               & Ours w/o Att.         & Ours                     \\ \hline
Straight     & 99 $\pm$ 2             & 92 $\pm$ 6          & \textbf{100} $\pm$ 0  & 99 $\pm$ 1               \\
One Turn     & \textbf{99} $\pm$ 2    & 83 $\pm$ 12         & 91 $\pm$ 8            & \textbf{99} $\pm$ 1      \\
Navigation   & 88 $\pm$ 7             & 68 $\pm$ 24         & 90 $\pm$ 9            & \textbf{96} $\pm$ 6      \\
Nav. Dyn.    & 86 $\pm$ 11            & 67 $\pm$ 21         & 88 $\pm$ 9            & \textbf{91} $\pm$ 9      \\ \hline
\end{tabular}
\end{center}
\caption{
Comparison of the success rate of the proposed approach to the state-of-the-art models on the original CARLA benchmark (CoRL2017) \cite{bib:carlasim2017}. Mean and standard deviation over three repetitions are reported (MT \cite{bib:rethinking2018} has two repetitions). For all methods, seed is not fixed at training time. Our approach emits competitive or better results than the previous state-of-the-art models of end-to-end driving.}
\label{tab:corl}
\end{table}

\begin{table*}
\begin{center}
\setlength\tabcolsep{5pt}
\begin{tabular}{lcccc|cccc}
\bhline{1pt}
             & \multicolumn{4}{c|}{Training conditions}                                            & \multicolumn{4}{c}{New weather}                                                          \\
Task         & MT                     & CILRS          & Ours w/o Att.   & Ours                    & MT                      & CILRS           & Ours w/o Att.    & Ours                      \\ \hline
Straight     & \textbf{0.0}$\pm$0.0   & 1.0$\pm$1.7    & 0.7$\pm$1.5     & 0.3$\pm$1.1             & 1.0$\pm$1.7             & 2.9$\pm$2.1     & 3.6$\pm$4.0      & \textbf{0.0}$\pm$0.0      \\
One Turn     & 0.3$\pm$0.7            & 1.1$\pm$0.9    & 0.8$\pm$1.2     & \textbf{0.0}$\pm$0.0    & \textbf{0.0}$\pm$0.0    & 0.6$\pm$1.3     & 1.1$\pm$1.1      & \textbf{0.0}$\pm$0.0      \\ 
Navigation   & 1.0$\pm$1.1            & 1.8$\pm$1.3    & 0.8$\pm$1.0     & \textbf{0.1}$\pm$0.3    & 0.6$\pm$0.6             & 0.6$\pm$0.6     & 0.6$\pm$0.9      & \textbf{0.4}$\pm$0.6      \\
Nav. Dyn.    & 0.6$\pm$0.6            & 1.1$\pm$1.3    & 0.6$\pm$1.1     & \textbf{0.5}$\pm$0.9    & 1.5$\pm$0.5             & 2.6$\pm$1.1     & 2.0$\pm$1.3      & \textbf{0.2}$\pm$0.4      \\ \hline
\bhline{1pt} \\ 
\vspace{-7mm} \\
\bhline{1pt}
             & \multicolumn{4}{c|}{New town}                                                     & \multicolumn{4}{c}{New town \& weather}                                                    \\
Task         & MT               & CILRS           & Ours w/o Att.      & Ours                    & MT                       & CILRS             & Ours w/o Att.      & Ours                   \\ \hline
Straight     & 15.9$\pm$10.1    & 8.3$\pm$4.5     & 10.3$\pm$3.5       & \textbf{5.4}$\pm$4.5    & 17.7$\pm$5.9             & 17.3$\pm$20.5     & 11.8$\pm$3.4       & \textbf{7.8}$\pm$6.5   \\
One Turn     & 11.1$\pm$5.3     & 11.3$\pm$4.8    & 8.2$\pm$3.9        & \textbf{4.1}$\pm$3.7    & 15.7$\pm$6.7             & 22.3$\pm$19.8     & 14.8$\pm$8.1       & \textbf{4.2}$\pm$3.3   \\ 
Navigation   & 9.7$\pm$6.4      & 12.0$\pm$6.7    & 12.9$\pm$2.8       & \textbf{8.7}$\pm$4.9    & 13.6$\pm$5.6             & 16.8$\pm$11.9     & 18.7$\pm$17.1      & \textbf{11.4}$\pm$6.9  \\
Nav. Dyn.    & 8.7$\pm$3.5      & 9.8$\pm$4.6     & 11.4$\pm$7.1       & \textbf{6.1}$\pm$3.6    & \textbf{9.5}$\pm$4.0     & 10.1$\pm$3.8      & 10.0$\pm$1.75      & 14.2$\pm$10.5          \\
\bhline{1pt}

\end{tabular}
 \caption{\label{tab:tl_infraction_corl}
 Traffic light infraction analysis on CoRL2017 benchmark \cite{bib:carlasim2017}. Percentage of times an agent crossed on red traffic lights is reported with mean and standard deviation over three repetitions except for MT \cite{bib:rethinking2018} that has two repetitions (lower is better). Our proposed approach records a smaller number of infractions under most tasks.
 Note that the MT method contains traffic light classifier in our implementation, not present in the original MT \cite{bib:rethinking2018}.
 }
\end{center}

\end{table*}

\begin{table}
\begin{center}
\begin{tabular}{lcc|cc}
\hline
         & \multicolumn{4}{c}{Training conditions (4 weathers, Town01)}                                    \\
Task     & MT                    & CILRS              & Ours w/o Att.             & Ours                   \\ \hline
Empty    & \textbf{100} $\pm$ 0  & 94 $\pm$ 6         & 97 $\pm$ 4                & 99 $\pm$ 1             \\
Regular  & 90 $\pm$ 3            & 87 $\pm$ 5         & \textbf{91} $\pm$ 5       & 88 $\pm$ 5             \\
Dense    & 46 $\pm$ 11           & 41 $\pm$ 6         & 45 $\pm$ 14               & \textbf{53} $\pm$ 11   \\ \hline\hline
         & \multicolumn{4}{c}{New weather (2 weathers, Town01)}                                            \\
Task     & MT                   & CILRS               & Ours w/o Att.             & Ours                   \\ \hline
Empty    & \textbf{98} $\pm$ 3  & 85 $\pm$ 9          & 94 $\pm$ 7                & 97 $\pm$ 4             \\
Regular  & 84 $\pm$ 3           & 76 $\pm$ 8          & 87 $\pm$ 10               & \textbf{93} $\pm$ 5    \\
Dense    & \textbf{45} $\pm$ 9  & 35 $\pm$ 6          & 42 $\pm$ 7                & 44 $\pm$ 12            \\ \hline\hline
         & \multicolumn{4}{c}{New town (4 weathers, Town02)}                                               \\
Task     & MT                   & CILRS               & Ours w/o Att.             & Ours                   \\ \hline
Empty    & 66 $\pm$ 14          & 57 $\pm$ 11         & 76 $\pm$ 11               & \textbf{90} $\pm$ 12   \\
Regular  & 50 $\pm$ 10          & 44 $\pm$ 11         & 62 $\pm$ 13               & \textbf{69} $\pm$ 11   \\
Dense    & 20 $\pm$ 4           & 17 $\pm$ 8          & 23 $\pm$ 5                & \textbf{37} $\pm$ 9    \\ \hline\hline
         & \multicolumn{4}{c}{New town \& weather (2 weathers, Town02)}                                    \\
Task     & MT                   & CILRS               & Ours w/o Att.             & Ours                   \\ \hline
Empty    & 57 $\pm$ 10          & 37 $\pm$ 11         & 71 $\pm$ 11               & \textbf{81} $\pm$ 11   \\
Regular  & 44 $\pm$ 7           & 30 $\pm$ 11         & 61 $\pm$ 8                & \textbf{67} $\pm$ 9    \\
Dense    & \textbf{24} $\pm$ 10 & 14 $\pm$ 5          & 22 $\pm$ 10               & 23 $\pm$ 5             \\ \hline
\end{tabular}
\end{center}
 \caption{\label{tab:NoCrash}
 Results on the NoCrash benchmark \cite{bib:codevilla2019} with mean and standard deviation over three runs except for MT \cite{bib:codevilla2019} that has two runs. For all methods, seed is not fixed when training.
 }

\end{table}

\begin{table*}
\begin{center}
\begin{tabular}{lcccc|cccc}
\bhline{1pt}
             & \multicolumn{4}{c|}{Training conditions}                                        & \multicolumn{4}{c}{New weather}                                                   \\
Task         & MT                   & CILRS        & Ours w/o Att.        & Ours               & MT                    & CILRS           & Ours w/o Att.  & Ours                   \\ \hline
Empty        & \textbf{0.7}$\pm$0.7 & 1.8$\pm$1.4  & 1.0$\pm$0.9          & 0.8$\pm$0.6        & \textbf{0.7}$\pm$0.8  & 3.0$\pm$1.4     & 1.2$\pm$1.1    & 1.8$\pm$1.2            \\
Regular      & \textbf{0.8}$\pm$0.7 & 1.7$\pm$1.5  & \textbf{0.8}$\pm$0.9 & 1.0$\pm$0.7        & \textbf{0.6}$\pm$0.6  & 3.3$\pm$1.1     & 1.2$\pm$1.1    & 1.9$\pm$1.4            \\
Dense        & 9.3$\pm$13.4         & 7.3$\pm$5.2  & \textbf{3.0}$\pm$2.1 & 10.7$\pm$19.1      & \textbf{4.4}$\pm$1.2  & 13.8$\pm$10.1   & 4.8$\pm$2.8    & 13.5$\pm$17.7          \\
\bhline{1pt} \\
\vspace{-7mm} \\
\bhline{1pt}
             & \multicolumn{4}{c|}{New town}                                                   & \multicolumn{4}{c}{New town \& weather}                                           \\
Task         & MT                    & CILRS         & Ours w/o Att.   & Ours                  & MT                    & CILRS          & Ours w/o Att.   & Ours                   \\ \hline
Empty        & 8.9$\pm$5.6           & 10.8$\pm$7.8  & 10.6$\pm$2.2    & \textbf{7.1}$\pm$4.1  & 15.8$\pm$5.2          & 16.5$\pm$13.8  & 21.1$\pm$14.7   & \textbf{15.5}$\pm$14.4 \\
Regular      & 6.5$\pm$3.9           & 7.2$\pm$4.2   & 7.1$\pm$1.9     & \textbf{5.1}$\pm$2.5  & \textbf{8.9}$\pm$3.7  & 10.8$\pm$4.5   & 9.4$\pm$2.4     & 13.2$\pm$17.1          \\
Dense        & \textbf{11.2}$\pm$6.2 & 12.1$\pm$8.8  & 12.8$\pm$8.7    & 16.3$\pm$19.1         & 8.1$\pm$3.6           & 8.9$\pm$3.4    & 21.8$\pm$19.0   & \textbf{3.1}$\pm$2.5   \\
\bhline{1pt}

\end{tabular}
 \caption{\label{tab:tl_infraction_nocrash}
 Traffic light infraction analysis on NoCrash benchmark \cite{bib:codevilla2019}. As in Table \ref{tab:tl_infraction_corl}, percentage of times an agent crossed on red traffic lights is reported with mean and standard deviation over three repetitions except for MT that has two repetitions (lower is better).
 As in Table \ref{tab:tl_infraction_corl}, MT contains traffic light classifier in our implementation, not present in the original MT \cite{bib:rethinking2018}.
 }
\end{center}
\end{table*}

\begin{figure*}
    \begin{center}
    \begin{tabular}{c}
        \begin{minipage}[c]{0.18\linewidth}
            \centering
            Input image
        \end{minipage}
        \begin{minipage}[c]{0.18\linewidth}
            \centering
            CILRS
        \end{minipage}
        \begin{minipage}[c]{0.18\linewidth}
            \centering
            MT
        \end{minipage}
        \begin{minipage}[c]{0.18\linewidth}
            \centering
            Ours w/o Att.
        \end{minipage}
        \begin{minipage}[c]{0.18\linewidth}
            \centering
            Ours
        \end{minipage}
    \end{tabular}
    
    Grad-CAM calculated on control prediction
    
    \begin{tabular}{c}
        \begin{minipage}[c]{0.18\linewidth}
            \centering
            \includegraphics[width=1.\columnwidth]{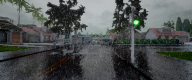}
            \label{fig:1}
        \end{minipage}
        \begin{minipage}[c]{0.18\linewidth}
            \centering
            \includegraphics[width=1.\columnwidth]{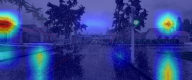}
            \label{fig:2}
        \end{minipage}
        \begin{minipage}[c]{0.18\linewidth}
            \centering
            \includegraphics[width=1.\columnwidth]{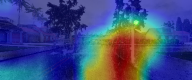}
            \label{fig:3}
        \end{minipage}
        \begin{minipage}[c]{0.18\linewidth}
            \centering
            \includegraphics[width=1.\columnwidth]{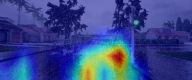}
            \label{fig:4}
        \end{minipage}
        \begin{minipage}[c]{0.18\linewidth}
            \centering
            \includegraphics[width=1.\columnwidth]{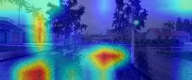}
            \label{fig:5}
        \end{minipage}
    \end{tabular} \\
    \vspace{-11pt}
    
    \begin{tabular}{c}
        \begin{minipage}[c]{0.18\linewidth}
            \centering
            \includegraphics[width=1.\columnwidth]{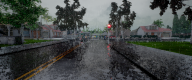}
            \label{fig:6}
        \end{minipage}
        \begin{minipage}[c]{0.18\linewidth}
            \centering
            \includegraphics[width=1.\columnwidth]{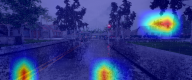}
            \label{fig:7}
        \end{minipage} 
        \begin{minipage}[c]{0.18\linewidth}
            \centering
            \includegraphics[width=1.\columnwidth]{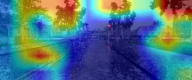}
            \label{fig:8}
        \end{minipage}
        \begin{minipage}[c]{0.18\linewidth}
            \centering
            \includegraphics[width=1.\columnwidth]{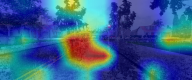}
            \label{fig:9}
        \end{minipage}
        \begin{minipage}[c]{0.18\linewidth}
            \centering
            \includegraphics[width=1.\columnwidth]{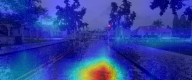}
            \label{fig:10}
        \end{minipage}
    \end{tabular} \\
    \vspace{-11pt}

    \begin{tabular}{c}
        \begin{minipage}[c]{0.18\linewidth}
            \centering
            \includegraphics[width=1.\columnwidth]{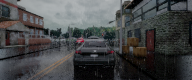}
            \label{fig:11}
        \end{minipage}
        \begin{minipage}[c]{0.18\linewidth}
            \centering
            \includegraphics[width=1.\columnwidth]{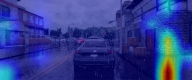}
            \label{fig:12}
        \end{minipage}
        \begin{minipage}[c]{0.18\linewidth}
            \centering
            \includegraphics[width=1.\columnwidth]{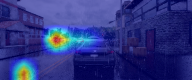}
            \label{fig:13}
        \end{minipage}
        \begin{minipage}[c]{0.18\linewidth}
            \centering
            \includegraphics[width=1.\columnwidth]{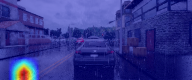}
            \label{fig:14}
        \end{minipage}
        \begin{minipage}[c]{0.18\linewidth}
            \centering
            \includegraphics[width=1.\columnwidth]{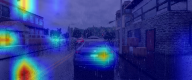}
            \label{fig:15}
        \end{minipage}
    \end{tabular} \\
    \vspace{-11pt}
    
    \begin{tabular}{c}
        \begin{minipage}[c]{0.18\linewidth}
            \centering
            \includegraphics[width=1.\columnwidth]{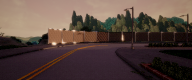}
            \label{fig:16}
        \end{minipage}
        \begin{minipage}[c]{0.18\linewidth}
            \centering
            \includegraphics[width=1.\columnwidth]{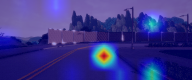}
            \label{fig:17}
        \end{minipage}
        \begin{minipage}[c]{0.18\linewidth}
            \centering
            \includegraphics[width=1.\columnwidth]{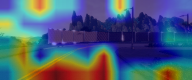}
            \label{fig:18}
        \end{minipage}
        \begin{minipage}[c]{0.18\linewidth}
            \centering
            \includegraphics[width=1.\columnwidth]{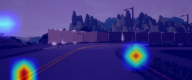}
            \label{fig:19}
        \end{minipage}
        \begin{minipage}[c]{0.18\linewidth}
            \centering
            \includegraphics[width=1.\columnwidth]{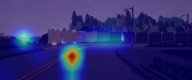}
            \label{fig:20}
        \end{minipage}
    \end{tabular} \\
    \vspace{-8pt}
    
    Grad-CAM calculated on traffic light state prediction
    
    \begin{tabular}{c}
        \begin{minipage}[c]{0.18\linewidth}
            \centering
            \includegraphics[width=1.\columnwidth]{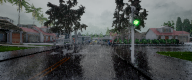}
            \label{fig:21}
        \end{minipage}
        \begin{minipage}[c]{0.18\linewidth}
            \centering
            \includegraphics[width=1.\columnwidth, height=0.416\columnwidth]{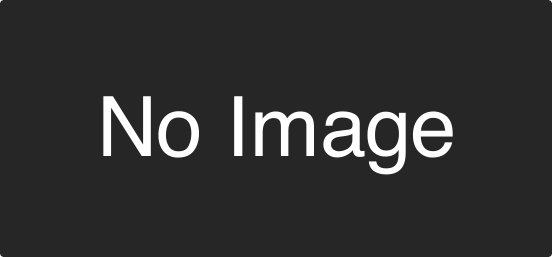}
            \label{fig:22}
        \end{minipage}
        \begin{minipage}[c]{0.18\linewidth}
            \centering
            \includegraphics[width=1.\columnwidth]{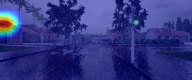}
            \label{fig:23}
        \end{minipage}
        \begin{minipage}[c]{0.18\linewidth}
            \centering
            \includegraphics[width=1.\columnwidth]{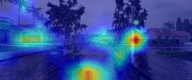}
            \label{fig:24}
        \end{minipage}
        \begin{minipage}[c]{0.18\linewidth}
            \centering
            \includegraphics[width=1.\columnwidth]{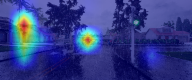}
            \label{fig:25}
        \end{minipage}
    \end{tabular} \\
    \vspace{-11pt}

    \begin{tabular}{c}
        \begin{minipage}[c]{0.18\linewidth}
            \centering
            \includegraphics[width=1.\columnwidth]{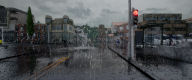}
            \label{fig:26}
        \end{minipage}
        \begin{minipage}[c]{0.18\linewidth}
            \centering
            \includegraphics[width=1.\columnwidth, height=0.416\columnwidth]{figs/no_image.pdf}
             \label{fig:27}
        \end{minipage}
        \begin{minipage}[c]{0.18\linewidth}
            \centering
            \includegraphics[width=1.\columnwidth]{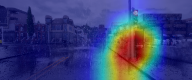}
            \label{fig:28}
        \end{minipage}
        \begin{minipage}[c]{0.18\linewidth}
            \centering
            \includegraphics[width=1.\columnwidth]{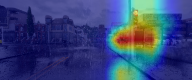}
            \label{fig:29}
        \end{minipage}
        \begin{minipage}[c]{0.18\linewidth}
            \centering
            \includegraphics[width=1.\columnwidth]{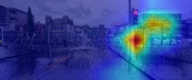}
            \label{fig:30}
        \end{minipage}
    \end{tabular} \\
 \vspace{-12pt}

 \caption{\label{fig:gradcam_results}
 Randomly picked Grad-CAM \cite{bib:GradCAM2017ICCV} visualization results. CILRS does not have traffic light prediction. Shown results on control prediction (top four rows) are summed over individual Grad-CAM results done on each control output.
 }
\end{center} 
\end{figure*}

\section{\label{sec:experiments} Experiments}

\subsection{\label{sec:training_setup} Training setup}

We implemented the proposed approach in TensorFlow 2.3.0 framework
\cite{bib:tensorflow-2015}.
In contrast to the  \cite{bib:rethinking2018}, we do not divide the
training process into stages.
We use a mini-batch size of $32$ and the Adam optimizer \cite{bib:adam2015}
in which the initial learning rate is set to $0.005$.
When training, we validate the presented model at the end of every epoch,
and then if the validation error of control does not record the best for
over five epochs, we divide the learning rate by $5$. The training lasts
for approximately two days on a single Nvidia GeForce GTX 1080Ti.
The resolution of all images is $384 \times 160$, including depth and
semantic segmentation images.
Without setting a random seed, we repeated all methods three times from
training to evaluation (except for MT, which was repeated only twice),
while the best seed out of five runs was selected in \cite{bib:codevilla2019}.
This may result in larger standard deviations in the results,
but the reproducibility may also be higher.
Note that we have come up with the above hyperparameter settings through experiments.

\subsection{Baselines}

We compare our proposed method to the two methods we build on:
the conditional imitation learning approach (CILRS) proposed
by Codevilla \etal \cite{bib:codevilla2019} and the original
multi-task learning approach (MT) proposed by Li \etal~\cite{bib:rethinking2018}
(see Section \ref{sec:method} for
detailed information). For a fair comparison, we implement
the two approaches with the same setup as ours using our dataset
and the same network module for all methods (just reflecting
the original idea of the papers),
so results are different from the original ones.
Also, note that MT contains a traffic light classifier that 
does not appear in the original implementation.

\subsection{\label{sec:experimental_setup} Experimental setup}

We evaluate our proposed method on the original CARLA benchmark named
CoRL2017 \cite{bib:carlasim2017} and on the NoCrash benchmark
\cite{bib:codevilla2019}. Both evaluate an agent in a
goal-directed navigation scenario whether or not the agent can accomplish
point-to-point navigation in a suburban setting. 
Each benchmark defines a fixed number of starting and goal points
in both Town01 and Town02, accompanied by a route planner that computes
a global path between starting and goal points using the A* algorithm
\cite{bib:A_Star}.
This planner sends the high-level command $\bm{c}$ at every frame considering
the location of an agent to indicate which direction it should proceed
from its current position. All evaluation trials are done in four
generalization contexts: training condition, new weather, new town,
and new town \& new weather. ``Training condition'' indicates that each
trial is done under four training weather conditions in Town01, which are
encountered when training, whereas ``new weather'' specifies two new
weather conditions that do not appear when training (see examples for the
weather conditions shown in Figure \ref{fig:data_examples}).
Also, as suggested by its name, ``new town'' indicates Town02 which
is unseen when training, and thus, ``new town \& new weather'' specifies
the most difficult full generalization test.

The CoRL2017 benchmark consists of four different driving conditions,
which are ``driving straight'', ``driving with one turn at an intersection'',
``full navigation with multiple turns at intersections'', and ``The same
full navigation but with dynamic obstacles''. A trial is considered successful
if an agent reaches within $2$ meters from the defined goal point.
Passing through a red traffic light is not counted as a failure.
Therefore, this benchmark is mainly designed for evaluating the ability
of lane-keeping and taking turns to navigate to the specified goal, but 
those high-level behaviors like handling dense traffic scenarios are not required.

We use NoCrash benchmark \cite{bib:codevilla2019} for more complex scenarios.
It measures the ability of agents to handle more complex
events in both towns under six weather conditions as in CoRL2017.
This benchmark defines $25$ goal-conditioned routes with
a difficulty level equivalent to the ``full navigation``
condition of the CoRL2017 benchmark, with three traffic
congestions:
1) empty town: no dynamic agents, 2) regular traffic: moderate
number of cars and pedestrians, 3) dense traffic: a large number
of pedestrians and heavy traffic. 

Unlike in CoRL2017, the NoCrash benchmark
considers an episode as a failure if a collision 
bigger than a predefined threshold occurs.
This makes the evaluation more similar to
the ones in the real-world setting where the
counted number of human interventions per
kilometer is a common metric and those human
interventions will put the vehicle back to the 
safe state
\cite{bib:codevilla2019}.
Additionally, this benchmark also measures the percentage of traffic
light violations, however, an infraction of crossing red traffic
light does not terminate a trial.

\subsection{\label{sec:benchmark_results}Evaluation results}

The results for the CoRL2017 benchmark are shown in Table \ref{tab:corl}, 
where we can see our implementation outperforms the baselines. Also,
both of our models outperform or are comparable with the current
state-of-the-art CILRS. Especially, although the performance of the
CILRS model decreases as the driving task gets harder in the ``new town''
and ``new town \& weather'' condition, the other models that have
multi-task knowledge can keep its performance to some extent,
and furthermore, our proposed approach stays at over 91 in all tasks. 
Table \ref{tab:tl_infraction_corl} reports the percentage of times an
agent crossed on red traffic lights in each driving condition on the
CoRL2017 benchmark. 
From this table, we can see that our proposed
approach is much less likely to experience the infraction of crossing
on red traffic lights in every condition. Thus, adding our task-specific
attention mechanism can enhance the capability of reacting to traffic lights.

The NoCrash benchmark results are reported in Table \ref{tab:NoCrash},
along with the traffic light violation rates reported in Table
\ref{tab:tl_infraction_nocrash}. Similar to the results on the first CoRL2017
benchmark, our proposed approach overall records the best numbers in the ``new town''.
We noticed, when we were visualizing the benchmarking runs,
there were some scenes where the agent got stuck every time
in those situations where it was exposed to strong direct
sunlight from the front covering huge regions of the input
image.
We believe the noisy numbers of traffic violations are a result
of visual difference between training and testing set 
(especially ``Wet Sunset'' and ``Soft Rain Sunset'' conditions in testing)
and the crowded traffic in the NoCrash benchmark.

Comparing MT with our proposed approach without attention, these two approaches
showed overall the same performance on the NoCrash benchmark, but occasionally
MT appeared to overfit the training town (also in CoRL2017). 
This gives us an intuition that learning the control task and scene representation
simultaneously allows the model to learn a more generalizable representation.
Interestingly, the MT approach outperformed the CILRS both
in the success rate of episodes and in handling traffic lights, which was the
opposite of the results reported in \cite{bib:codevilla2019}.

\subsection{Grad-CAM visualizations}
For the qualitative analysis, we use the Grad-CAM visualization
\cite{bib:GradCAM2017ICCV} shown in Figure \ref{fig:gradcam_results}
for all methods using randomly picked images from training conditions. 
Grad-CAM allows us to study what parts of the input play important role
in deciding the output.
Firstly, we visualized salient parts that contributed to control outputs.
We can see that Grad-CAM attention maps of all methods may cover meaningful
regions like the center line or the border between road and sidewalk.
When taking a close look, CILRS might spot irrelevant points and MT appeared
to be distracted into large regions, while our proposed approach looks at 
narrow yet important points like the center of the road or leading vehicle.
Secondly, we also visualized attention maps based on traffic light prediction.
MT and our proposed approach without attention widely capture the traffic
light including the pole when it is red, whereas our proposed approach attentively
looks at it but the red color point is not included.

\section{Conclusion}
In this work, we have presented a novel multi-task
attention-aware network model for end-to-end autonomous
driving in conditional imitation learning framework \cite{bib:cil2018}. 
This model uses two types of attention paths to generate
task-specific feature maps fed into each task-specific module.
To verify its effectiveness, we conducted experiments on two
CARLA benchmarks \cite{bib:codevilla2019, bib:carlasim2017}
and quantitatively showed that the attention improves the learned
control policy including the ability to handle traffic lights, 
outperforming the baseline methods.
By visualizing the attended regions using Grad-CAM,
we find that our model attends to correct points or objects
of interest (e.g. center lines, other cars) when making control decisions.

One future direction would be to include temporal information as well, e.g. by using videos.
This way, the end-to-end driving model can more precisely
capture the visual inputs by distinguishing dynamic
and static objects.
Also, it would be interesting to see if the proposed
method can transfer to a real-world data such as BDD100K dataset \cite{bib:bdd100k},
after which it would be possible to show how a real car would behave
when controlled by our approach.

{\small
\bibliographystyle{ieee_fullname}
\bibliography{egbib}
}

\end{document}